\documentclass[10pt,twocolumn,letterpaper]{article}

\usepackage[pagenumbers]{cvpr} 
\usepackage{tabularx}
\usepackage{algorithm}
\usepackage{algorithmic}
\usepackage{amsmath} 
\usepackage{arydshln}
\definecolor{cvprblue}{rgb}{0.21,0.49,0.74}
\usepackage[pagebackref,breaklinks,colorlinks,allcolors=cvprblue]{hyperref}

\def\paperID{****} 
\def\confName{CVPR}
\def\confYear{2026}

\title{Active Intelligence in Video Avatars via Closed-loop World Modeling}

\author{
    Xuanhua He$^{1,*}$ \quad
    Tianyu Yang$^{2,\dagger}$ \quad
    Ke Cao$^{3}$ \quad
    Ruiqi Wu$^{2}$ \quad
    Cheng Meng$^{2}$ \\
    Yong Zhang$^{2,\ddagger}$ \quad
    Zhuoliang Kang$^{2}$ \quad
    Xiaoming Wei$^{2}$ \quad
    Qifeng Chen$^{1,\dagger}$
    \\[0.5cm] 
    $^{1}$The Hong Kong University of Science and Technology \quad
    $^{2}$Meituan \\  
    $^{3}$University of Science and Technology of China\\
    \url{https://xuanhuahe.github.io/ORCA/}
}

\begin{document}
\maketitle
\begingroup
\renewcommand\thefootnote{} 
\footnotetext{$^*$Work done during an internship at Meituan. $^\dagger$Corresponding authors. $^\ddagger$Project leader.}
\endgroup
\begin{figure*}
    \centering
    \includegraphics[width=\linewidth]{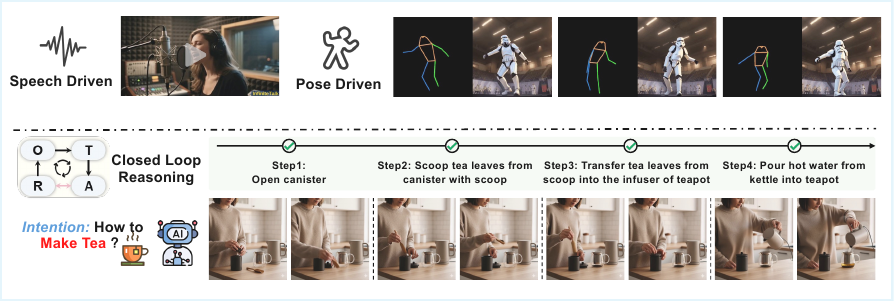}
    \put(-320, 101){\small{(a) Passive-condition-driven methods.}}
    \put(-335, 0){\small{(b) Online reasoning and cognitive architecture.}}
    \vspace{-0.2cm}
    \caption{Comparison of video avatar generation approaches.  (a) Speech-driven and pose-driven methods produce passive motions with limited semantic understanding. In contrast, (b) our \textbf{O}nline \textbf{R}easoning and \textbf{C}ognitive \textbf{A}rchitecture (ORCA) enables complex, multi-step task execution through OTAR (Observe-Think-Act-Reflect) closed-loop reasoning.}
    \label{fig:firstfig}
\end{figure*}
\begin{abstract}
Current video avatar generation methods excel at identity preservation and motion alignment but lack genuine agency—they cannot autonomously pursue long-term goals through adaptive environmental interaction. We address this by introducing L-IVA (Long-horizon Interactive Visual Avatar), a task and benchmark for evaluating goal-directed planning in stochastic generative environments, and ORCA (Online Reasoning and Cognitive Architecture), the first framework enabling active intelligence in video avatars. ORCA embodies Internal World Model (IWM) capabilities through two key innovations: (1) a closed-loop OTAR cycle (Observe-Think-Act-Reflect) that maintains robust state tracking under generative uncertainty by continuously verifying predicted outcomes against actual generations, and (2) a hierarchical dual-system architecture where System 2 performs strategic reasoning with state prediction while System 1 translates abstract plans into precise, model-specific action captions. By formulating avatar control as a POMDP and implementing continuous belief updating with outcome verification, ORCA enables autonomous multi-step task completion in open-domain scenarios. Extensive experiments demonstrate that ORCA significantly outperforms open-loop and non-reflective baselines in task success rate and behavioral coherence, validating our IWM-inspired design for advancing video avatar intelligence from passive animation to active, goal-oriented behavior. 
\end{abstract}    
\section{Introduction}
\label{sec:intro}
The generation and control of video avatars~\cite{lin2025omnihuman,jiang2025omnihuman,wang2025interacthuman} represents a 
frontier in computer vision, aiming to create virtual agents that 
exhibit human-like intelligence in their actions.
Recent methods enable the generation of full-body~\cite{Gan2025OmniAvatarEA,Hu2025AnimateA2} avatars that maintain high fidelity to a reference identity, conditioned on driving signals such as text, speech~\cite{Tian2024EMOEP}, or predefined pose sequences~\cite{Ma2023FollowYP}. For long video generation, these models typically operate in a chunk-level autoregressive manner, where each new video segment is generated conditioned on the final frames of the previous one~\cite{Yang2025InfiniteTalkAV,lin2025omnihuman,jiang2025omnihuman}.
Despite this progress in identity preservation and alignment with driving signals, the avatars generated by current methods still lack genuine agency; they can execute predefined actions or follow simple commands, but cannot autonomously pursue long-term goals through multi-step planning and adaptive interaction with their environment.  This gap from passive animation to active, goal-oriented behavior limits the application of video avatars in broader and more dynamic scenarios, such as virtual human livestreaming or autonomous product hosting.

This limitation raises a question: How can we transition from passive animation to active, goal-oriented intelligence?
To achieve this, an avatar must: (1) maintain understanding of task progress from incomplete visual observations (the agent only sees generated clips), (2) predict how actions affect future states, and (3) plan coherent action sequences toward long-term goals. 
This decision-making ability requires an \textbf{internal representation that synthesizes observation history to estimate the true world state}.
In cognitive science and control theory, such representations are formalized as \textit{Internal World Models (IWMs)}~\cite{diester2024internal}, which enable agents to: (i) estimate states from observations and predict future states, and (ii) simulate action outcomes for deliberate planning~\cite{wang2025vagen}. 
Formally, the setting of decision-making under partial observability with goal-directed planning is characterized as a Partially Observable Markov Decision Process (POMDP)~\cite{spaan2012partially}. Recent work in multi-turn VLM agents~\cite{wang2025vagen} and biological intelligence~\cite{diester2024internal} confirms that explicit world modeling is essential for robust long-horizon planning. For video avatars, IWMs are particularly critical as the agent must continuously track what has been accomplished and predict how each generated clip advances toward the goal, and the state of the environment which can't be directly seen from video.

However, realizing IWMs for generative video avatars presents fundamentally different challenges than in robotics or embodied AI, where agents interact with deterministic physical environments. We identify two core challenges:
\noindent\textbf{State Estimation and Tracking under Generative Uncertainty.} 
Traditional IWMs assume a fixed world model where repeated actions yield consistent outcomes. In contrast, video generation is inherently stochastic, the same action specification can produce diverse visual outcomes due to the probabilistic nature of I2V models. This creates a challenge in maintaining internal state. Without verifying generated outcomes against intended states, the agent's internal state may lead to errors in long-horizon planning. Unlike physical robots that can rely on sensors, video avatars must infer state solely from their own generated clips under partial observability. This necessitates a closed-loop mechanism that continuously grounds states through outcome verification.

\noindent\textbf{Planning in Open-domain Action Space.}
Unlike robotics with bounded action spaces (joint angles), video avatar actions are semantic and open-domain without predefined primitives. Simple action textual descriptions alone (``pick up the red cup") leave visual details unspecified, causing diverse, often incorrect generations. This demands hierarchical planning, not only decide the next action, but also translating this action into detailed, model-specific control signal to precise control of avatars.


To systematically evaluate active intelligence in video avatars, we propose 
\textbf{L-IVA (Long-horizon Interactive Visual Avatar) task and benchmark} for autonomous goal completion through multi-step environmental interactions. Unlike video generation benchmark evaluating single-clip aesthetics, L-IVA tests goal-directed planning in stochastic generative environments. The task is formulated as \textit{chunk-level autoregressive generation}: given an initial scene and high-level goal (e.g., ``host a product demo"), agents generate sequential video clips depicting coherent task completion via meaningful object interactions.

To succeed on L-IVA, we propose ORCA (Online Reasoning and Cognitive Architecture), which embodies IWM capabilities in this specific domain through two designs, grounded in established theories from control theory: 
To address Challenge 1's uncertainty and correctly update the internal state, OCRA works in an \textit{Observe-Think-Act-Reflect} (OTAR) loop.
The agent observe clips and history state to estimate current state and update the internal belief, think to plan sub-goals, and reflect to verify outcomes and trigger corrections. 
This prevents error accumulation in stochastic generation.
Second, to address challenge 2's action specification problem, we implement two specialized VLM modules inspired by dual-process theory~\cite{evans2013dual}. \textbf{System 2} works in \textit{Observe-Think-Reflect} stage, it evaluates progress, plans sub-goals, and predict next state, addressing open-domain planning through reasoning. \textbf{System 1}  operates during \textit{Act} stage, it translates abstract sub-goals into detailed action captions which are specially designed for specific I2V model for execution.
Together, these innovations enable autonomous behavior where System 2 maintains strategic coherence over long horizons and System 1 ensures execution precision. The OTAR cycle provides robustness in probabilistic generative environments. As illustrated in Figure~\ref{fig:firstfig}, our method enables multi-step complex task generation, in contrast to passive animation methods such as speech-driven or pose-driven approaches.
Our contributions can be summarized as follows:
\begin{itemize}
    \item We introduce L-IVA benchmark for evaluating video avatars' capability for autonomous, long-horizon task completion in interactive scenarios.
    \item We propose ORCA, the first framework enabling active intelligence in generative video avatars through: (1) a closed-loop OTAR cycle for robust state tracking and execution in stochastic generative worlds, and (2) a dual-system hierarchical architecture for open-domain planning with precise action grounding.
    \item Through extensive experiments on L-IVA, we demonstrate that ORCA outperforms open-loop and non-reflective baselines in task success rate and behavioral coherence, validating the effectiveness of our IWM-inspired design for video avatar intelligence.
\end{itemize}

\section{Related Work}
\label{sec:RelatedWork}
\subsection{Video Avatar Model}
The creation of controllable video avatars has fallen into two categories: generating motion~\cite{Ma2023FollowYP,ma2025sayanything,prajwal2020lip,
lin2025omnihuman,jiang2025omnihuman} for a given human image and maintaining consistent identity~\cite{he2024id,huang2024consistentid,zhong2025concat,wang2025dynamicface} given a facial image. Audio-driven methods translate speech into motion, evolving from two-stage pipelines using 3D meshes to end-to-end models~\cite{Yang2025InfiniteTalkAV,jiang2025omnihuman} that directly synthesize video. Identity-consistent text to video generation preserves appearance from reference images using textual prompts~\cite{he2024id,zhong2025concat,Tu2024StableAnimatorHI}. Recent works like InterActHuman~\cite{wang2025interacthuman} merge these streams, driving consistent identities with audio and text. However, these approaches frame generation as reactive signal processing—motion as audio response, identity as feature fusion—lacking goal-oriented planning. We introduce cognitive reasoning that plans actions from long-term goals before synthesis, enabling purposeful behavior beyond reactive generation.


\subsection{Agent for Video Generation}
Recent agent-based frameworks tackle complex video creation through task decomposition. Multi-agent systems like DreamFactory~\cite{xie2024dreamfactory}, 
StoryAgent~\cite{hu2024storyagent}, and Mora~\cite{yuan2024mora} employ specialized agents for story design, storyboard generation, and synthesis to ensure narrative coherence across scenes. VideoGen-of-Thought~\cite{zheng2024videogen} decomposes sentences into multi-shot storylines. A second class introduces iterative refinement: 
VISTA~\cite{long2025vista} and GENMAC~\cite{huang2024genmac} employ generation-critique cycles to enhance prompts for improving clip quality.
Despite these advances, prior agents focus on perfecting single clips
through feedback correcting deviations from initial prompts. In contrast, our framework maintains goal-directed behavior across long horizons, where each clip is a step in an evolving interaction rather than an endpoint to refine.

\subsection{World Models for Agentic Planning}
Internal world models enable agents to maintain an internal belief while estimating states from observations and predict action outcomes to make decisions under partial observability~\cite{diester2024internal}. In embodied AI, world models support planning through learned forward dynamics~\cite{cen2025worldvla}. Recent work extends this concept to VLM agents for game playing~\cite{gao2025vision,wang2025vagen,chen2025internalizing} and navigation through RL training. However, these methods assume low-variance environments where repeated actions yield consistent outcomes, which differ from our work in a generative environment.

\section{Method}
\label{sec:Method}
In this section, we first formally define our proposed L-IVA task. We then present the ORCA architecture. 
\begin{figure*}
    \centering
    \includegraphics[width=\linewidth]{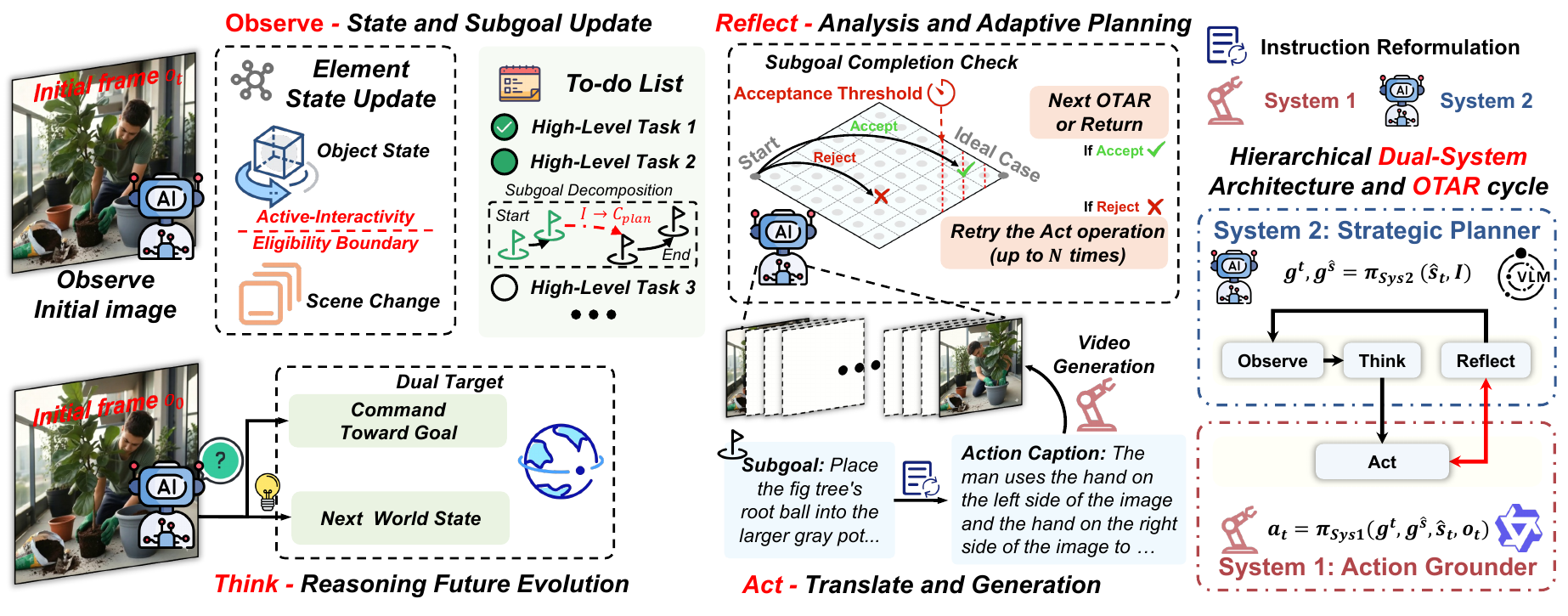}
    \caption{Overview of the ORCA framework. ORCA operates through a closed-loop OTAR cycle: Observe updates internal world state from generated clips, Think (System 2) decomposes tasks and predict next state, Act (System 1) translates subgoals into action captions for I2V generation, and Reflect verifies completion to accept/reject outcomes. This hierarchical dual-system architecture enables robust long-horizon task execution through continuous state tracking and adaptive replanning.}
    \label{fig:mainfig}
\end{figure*}
\subsection{L-IVA Task as POMDP}
\label{task_define}
We formulate a new task, which we term Long-horizon, Interactive Visual Avatar (L-IVA). Unlike traditional avatar animation tasks driven by predefined signals, L-IVA requires an agent to autonomously achieve high-level intentions through sequential interactions with a generative environment, treating an Image-to-Video (I2V) model as its world 
simulator.
In this task, the agent cannot  obtain the true world state directly from generated video frames as the frame only contains partial information, it must infer state from observation based on its knowledge. Moreover, identical actions yield diverse outcomes due to the stochastic I2V model. This combination of hidden states and generative stochasticity necessitates the POMDP framework~\cite{spaan2012partially}.
Formally, L-IVA is defined by the tuple 
$(\mathcal{S}, \mathcal{A}, \mathcal{T}, \mathcal{R}, \Omega, \mathcal{O})$:
The true state of the world, $s_t \in \mathcal{S}$, is a latent, unobservable representation encompassing the complete properties of the avatar and its environment. The agent interacts with this world through an open-ended action space $\mathcal{A}$ composed of natural language captions. Each action $a_t \in \mathcal{A}$ is a high-level command. 
The agent perceives the world through observations $o_t \in \Omega$—
sequences of video frames generated by the I2V model. The observation 
function $\mathcal{O}(o_t|s_t)$ provides only a partial view of the 
true state $s_t$.
The true state transition $\mathcal{T}(s_{t+1}|s_t, a_t)$ is implicit 
and hidden. What is observable is the stochastic observation 
defined by the pre-trained I2V model: $o_{t+1} \sim G_\theta(o_t, a_t)$. 
For a given high-level intention $I$, the reward $\mathcal{R}$ is sparse and terminal, yielding a value of 1 only if the final state of the trajectory satisfies the intention. Since true states $s_t$ are hidden, the agent cannot condition its policy directly on states. Instead, it maintains an internal \textit{belief state} $\hat{s}_t$, a state with history information and updated via observation. The agent must execute a policy $\pi(a_t | \hat{s}_t)$, conditioned on the belief state that maximizes the probability of task success.
\subsection{ORCA: IWM-inspired Architecture}
\subsubsection{Overview}
To succeed in L-IVA, we propose ORCA (Online 
Reasoning and Cognitive Architecture) realizes IWM capabilities through two innovations addressing the unique challenges of generative environments, as shown in Figure~\ref{fig:mainfig}.
ORCA operates in an \textit{Observe-Think-Act-Reflect} loop where predicted states are continuously verified against actual outcomes, triggering re-generation when mismatched. This closed-loop design prevents belief corruption caused by the randomness of the I2V model, where the model generated different results given the same caption.
Open-domain actions require both high-level strategic planning and low-level execution grounding, which operate at different levels of abstraction. ORCA adopts a dual-system architecture to address this: System 2 performs reasoning, while System 1 translates these plans into precise, I2V-compatible captions. This separation ensures both strategic coherence and execution fidelity.
Both designs leverage pre-trained VLMs through structured prompting, requiring no task-specific training. We detail each component below.

\subsubsection{Hierarchical Dual-System Architecture}
Effectively controlling a generative avatar requires balancing two demands at different levels of abstraction: high-level strategic reasoning for open-domain actions, and low-level, precise prompt to constrain the stochastic I2V model. A monolithic system struggles to manage this, since reliable generative control is often model-specific. To address this, ORCA adopts a Hierarchical Dual-System Architecture inspired by dual-process theory~\cite{evans2013dual}. This design decouples planning from execution: \textit{System 2} performs reasoning, while \textit{System 1} translates these abstract commands into detailed, I2V-model specific captions.


\noindent\textbf{System 2 for strategic reasoning} System 2 maintains the IWM's high-level belief state $\hat{s}_t$ and performs strategic reasoning. In the Think stage:
\begin{equation}
    g_{t},g_{\hat{s}} = \pi_{\text{Sys2}}(\hat{s}_t,
I)
\end{equation} where $\hat{s}_t$ is current belief state, containing the environment information, history $h_t$ and task checklist $\mathcal{C}$, and $g_{t},g_{\hat{s}}$ represent the textual command and the predicted next state respectively.

System 2's strategic focus enables it to leverage pre-trained VLMs' 
broad world knowledge for compositional reasoning over open-domain 
scenarios without being constrained by the specific formatting 
requirements of generative models. It operates in multiple OTAR stages: analyzing observations for state estimation and update, selecting sub-goals and predicting outcomes, and verifying results.

\noindent\textbf{System 1 for action grounding}
Operating in the Act stage, System 1 serves for grounding System 2's abstract plan into a concrete control signal. Precise avatar control is highly sensitive to the prompt and different I2V models require different prompt format; therefore, System 1's core task is to translate the multi-modal intention ($g_t, g_{\hat{s}}$) into a detailed action caption $a_t$ tailored for the specific I2V model $G_\theta$. This process leverages extensive prompt engineering to ensure high-fidelity translation. The grounding policy is defined as:
\begin{equation}
    a_t = \pi_{\text{Sys1}}(g_t, g_{\hat{s}}, o_t, \hat{s}_t) 
\end{equation}
where $a_t$ is the final, executable caption. Further details on our prompt engineering strategies are provided in the Appendix section~\ref{sec:prompts}.

\subsubsection{Closed-loop OTAR Cycle}
Open-loop plans fail easily in the L-IVA task, as even minor execution errors accumulate. A conventional Observe-Think-Act cycle is also insufficient as results from I2V model may significantly differ from the agent's intention. A failed generation can produce erroneous states that are difficult to recover from in subsequent steps, integrating these outcomes would corrupt the agent's internal belief.
ORCA addresses this through an \textbf{Observe-Think-Act-Reflect (OTAR)} cycle, where the Reflect stage continuously verifies outcomes against predictions before belief updates. This closed-loop design prevents belief corruption from stochastic generation. Algorithm~\ref{alg:orca_main} presents the complete procedure, which we detail stage-by-stage below.

\noindent\textbf{Initialization.} Given intention $I$ and initial scene $o_0$, System 2 initializes the belief state $\hat{s}_0 = (s_{\text{scene}}, C_{\text{plan}}, h_{\emptyset})$. $s_{\text{scene}}$ represents interactive objects and properties in $o_0$. $C_{\text{plan}}$ decomposes $I$ into a plan of sub-goals defined upon these objects. $h_{\emptyset}$ is the empty interaction history.

\noindent\textbf{Observe} At turn $t$, System 2 updates the belief state from the latest clip:
\begin{equation}
    \hat{s}_t = f_{\text{observe}}(o_t, \hat{s}_{t-1})
\end{equation}
producing structured observations of scene changes, updated object states, 
and completed sub-goals, maintaining the agent's understanding of task progress.

\noindent\textbf{Thinking}
System 2 plans the next action based on current belief 
$\hat{s}_t$, intention $I$, and current observation $o_t$:
\begin{equation}
    g_t, g_{\hat{s}} = f_{\text{think}}(\hat{s}_t, I, o_t)
\end{equation}
where $g_t$ is the command toward finishing the next goal and $g_{\hat{s}}$ is a detailed structural description of the predicted outcome state. 

\noindent\textbf{Action}
System 1 translates the abstract sub-goal $g_t$ into a detailed action caption $a_t$ tailored to the I2V model, then generates:
\begin{equation}
    v_{t+1} \sim G_\theta(o_t, a_t)
\end{equation}
where detailed text conditioning $a_t$ ensures generation fidelity. 

\noindent\textbf{Reflect}
System 2 verifies whether the generated outcome matches 
the predict states:
\begin{equation}
    \delta_t, \text{analysis} = f_{\text{reflect}}(o_{t+1}, g_t, g_{\hat{s}})
\end{equation}
producing $\delta_t \in \{\text{accept}, \text{reject}\}$. 
$o_{t+1}$ is the sampled frames from $v_{t+1}$.
If accepted, the agent proceeds to turn $t+1$ to update beliefs. If rejected, System 2 analyzes the failure and either retries with a revised action $a_t^{\text{new}} = f_{\text{revise}}(a_t, o_{t+1}, \text{analysis})$ 
(up to $N_{\text{retry}}$ attempts) or adaptively re-plans for the next iteration. This prevents belief corruption from failed generations.The cycle continues until $\hat{s}_t.\mathcal{C}_{\text{remaining}} = \emptyset$, returning video sequence $V = [v_1, ..., v_T]$.

By continuously verifying predictions and triggering corrections, ORCA maintains accurate beliefs despite generative uncertainty, enabling robust long-horizon task completion.
\begin{algorithm}[t]
\caption{ORCA: Closed-loop OTAR Execution}
\label{alg:orca_main}
\begin{algorithmic}[1]
\REQUIRE Initial observation $o_0$, Intention $I$, I2V model $G_\theta$, Max retries $N_{\text{retry}}$
\ENSURE Video sequence $V = [v_1, ..., v_T]$
\STATE $\hat{s}_0 \gets \text{Initialize}(o_0, I)$ \COMMENT{Scene analysis + plan decomposition}
\STATE $t \gets 0$, $V \gets []$
\WHILE{$\hat{s}_t.\mathcal{C}_{\text{remaining}} \neq \emptyset$}
    \STATE \textbf{// Observe Stage}
    \STATE $\hat{s}_t \gets f_{\text{observe}}(o_t, \hat{s}_{t-1})$
    
    \STATE \textbf{// Think Stage}
    \STATE $g_t, g_{\hat{s}} \gets f_{\text{think}}(\hat{s}_t, I, o_t)$    
    \STATE \textbf{// Act Stage}
    \STATE $a_t \gets \pi_{\text{Sys1}}(g_t,g_{\hat{s}}, o_t, \hat{s}_t)$
    \STATE $v_{t+1} \sim G_\theta(o_t, a_t)$
    \STATE $o_{t+1} \gets \text{SampleFrames}(v_{t+1})$
    
    \STATE \textbf{// Reflect Stage}
    \STATE $\text{retry\_count} \gets 0$
    \REPEAT
        \STATE $\delta_t, \text{analysis} \gets f_{\text{reflect}}(o_{t+1}, g_t, g_{\hat{s}})$
        \IF{$\delta_t = \text{reject}$ AND $\text{retry\_count} < N_{\text{retry}}$}
            \STATE $a_t \gets f_{\text{revise}}(a_t, o_{t+1}, \text{analysis})$
            \STATE $v_{t+1} \sim G_\theta(o_t, a_t)$
            \STATE $o_{t+1} \gets \text{SampleFrames}(v_{t+1})$
            \STATE $\text{retry\_count} \gets \text{retry\_count} + 1$
        \ENDIF
    \UNTIL{$\delta_t = \text{accept}$ OR $\text{retry\_count} \geq N_{\text{retry}}$}
    
    \IF{$\delta_t = \text{accept}$}
        \STATE $V.\text{append}(v_{t+1})$, $t \gets t+1$
    \ELSE
        \STATE Re-plan from current state \COMMENT{Adaptive recovery}
    \ENDIF
\ENDWHILE
\RETURN V
\end{algorithmic}
\end{algorithm}

\section{L-IVA Benchmark}
\label{sec:L-EVA Benchmark}
\subsection{Overview}
\begin{figure}
    \centering
    \includegraphics[width=\linewidth]{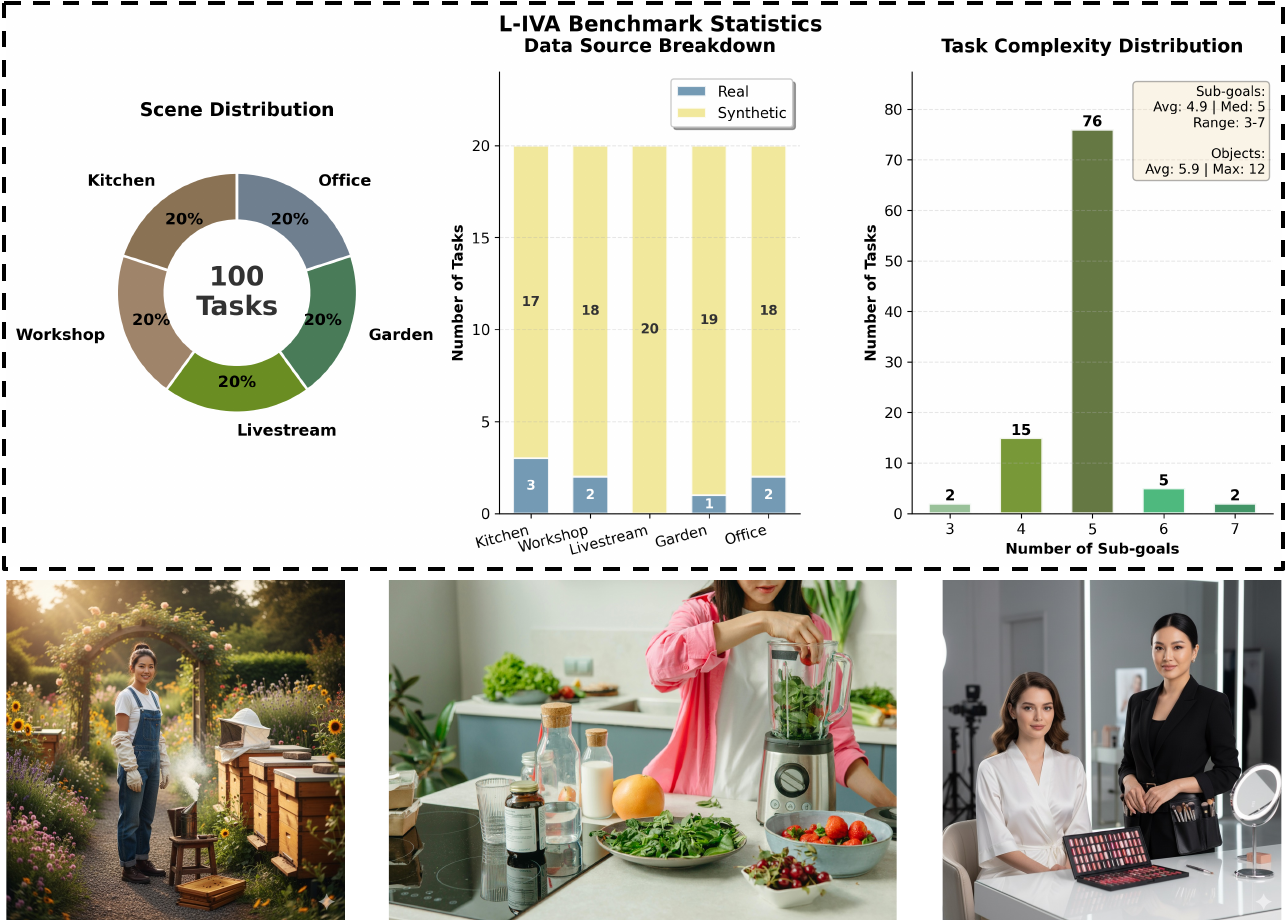}
\caption{\textbf{L-IVA Benchmark Overview.} \textit{Top}: Statistical analysis showing (left) balanced scene distribution across 5 categories, (center) data source composition with 92 synthetic and 8 real images, and (right) task complexity distribution averaging 5.0 sub-goals per task. \textit{Bottom}: Representative scenes from our benchmark including Garden, Kitchen, and livestream scenarios, demonstrating diverse real-world settings requiring multi-step object interactions.}
    \label{fig:dataset}
\end{figure}

While Section~\ref{task_define} formally defined the task of achieving active agency as a POMDP, empirically evaluating this capability requires a new benchmark. Existing frameworks are inadequate, video generation benchmarks evaluate per-clip aesthetics rather than long-horizon, goal-directed task completion.
We introduce the L-IVA Benchmark comprising 100 tasks across 5 real-world scenario categories. Notably, each category includes 5 two-person collaborative tasks. Each task requires 3-8 interaction steps with more than three objects, testing autonomous planning, environmental reasoning, and precise control. Tasks use fixed-viewpoint, single-room 
settings to avoid spatial inconsistencies in current I2V models. Figure~\ref{fig:dataset} summarizes task statistics; Appendix provides detailed specifications in section~\ref{sec:benchmark_details}.
\subsection{Dataset Construction}
Our benchmark scenes consist of AI-generated and real images. Each task is annotated with: (1) an {object inventory} (names, positions, initial states); (2) a high-level {natural language intention}; and (3) a {reference action sequence} as a ground-truth solution. The evaluation focuses on goal achievement over rigid imitation, measuring successful task completion and accepting alternative valid action orderings, not exact trajectory matching.
\subsection{Evaluation Protocol and Metrics}
We evaluate agents on L-IVA using VLM-based judgments and human annotations. Our primary metric is the {Task Success Rate (TSR)}, measuring sub-goal progress. TSR is:
\begin{equation}
    \text{TSR} = \frac{1}{N}\sum_{i=1}^{N}\frac{k_i}{M_i}
\end{equation}
The final score is weighted by the ratio of completed sub-goals ($k_i$) to total sub-goals ($M_i$), where sub-goal completion is manually verified by human annotators. To evaluate overall quality, we employ Best-Worst Scaling (BWS)~\cite{louviere2015best} to derive robust human preference rankings. Additionally, we introduce two diagnostic metrics, which are detailed in Section~\ref{pps}: (1) a human-rated {Physical Plausibility Score} to assess object permanence and spatial consistency; and (2) a VLM-based {Action Fidelity Score} to quantify the semantic alignment between commands and video clips ($a_t, v_t$). Together, this hybrid human-VLM evaluation framework provides a comprehensive assessment of goal-oriented success and execution reliability.
\section{Experiments}
\label{sec:exp}
\begin{figure*}
    \centering
    \includegraphics[width=\linewidth]{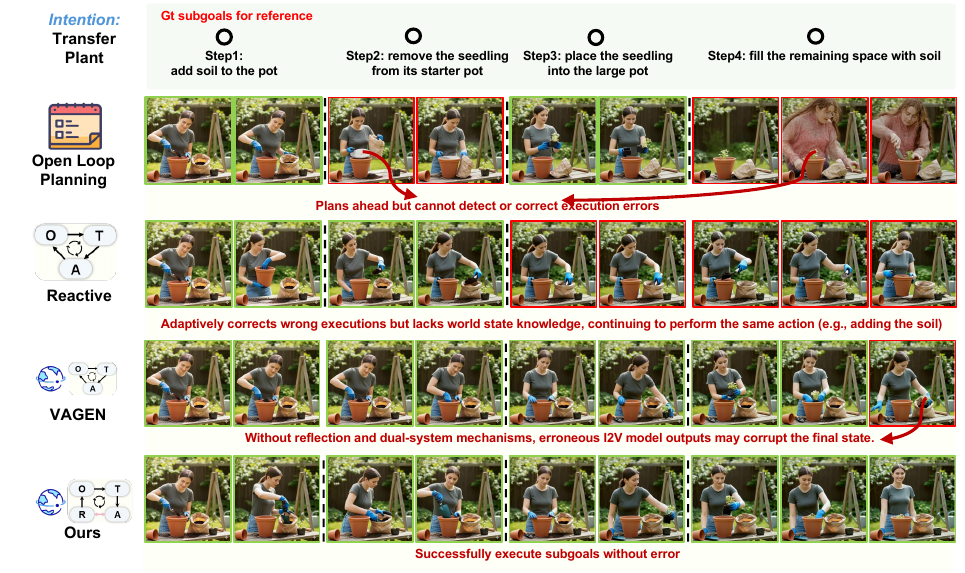}
    \caption{\textbf{Qualitative Comparison on Transfer Plant Task.} 
We compare four methods on long-horizon video generation. 
\textit{Top:} Ground truth subgoals for reference. 
\textcolor{red}{Red boxes} indicate execution failures or error accumulation.
Open-Loop planner cannot detect execution errors.
Reactive agent lacks world state knowledge, leading to repetitive actions.
VAGEN's I2V errors corrupt the final state without reflection.
\textbf{ORCA (Ours)} successfully completes all subgoals with consistent execution quality.
}
    \label{fig:comapfig}
\end{figure*}
\begin{table*}[!ht]
\caption{Main Results on L-IVA Benchmark. All metrics are evaluated per scenario. (a) Task completion metrics. (b) Video quality and human preference. Best in \textbf{bold}.}
\label{tab:main_results}
\centering

\vspace{0.3em}
\textbf{(a) Task Success Rate (\%) and Execution Quality}
\vspace{0.2em}

\small
\setlength{\tabcolsep}{3.5pt}
\begin{tabular}{l|ccccc|c||ccccc|c||ccccc|c}
\toprule
& \multicolumn{6}{c||}{Task Success Rate (\%) $\uparrow$} & \multicolumn{6}{c||}{Physical Plausibility (1-5) $\uparrow$} & \multicolumn{6}{c}{Action Fidelity (0-1) $\uparrow$} \\
\cmidrule{2-7} \cmidrule{8-13} \cmidrule{14-19}
Method & Kit. & Live. & Work. & Gard. & Off. & Avg & Kit. & Live. & Work. & Gard. & Off. & Avg & Kit. & Live. & Work. & Gard. & Off. & Avg \\
\midrule
Reactive & 56.7& 58.7& 45.8& 55.0& 38.1& 50.9& 3.47& 3.05& 3.13& 3.08& 2.80& 3.11& 0.53& 0.56& 0.41& 0.61& 0.63& 0.55\\
Open-Loop &  72.3& \textbf{72.3}& 72.7& 46.2&     47.9& 62.3& \textbf{3.57}& 3.50& 3.27& 2.92& 2.60& 3.17& 0.57& \textbf{0.72}& 0.53& \textbf{0.64}& 0.65& 0.62\\
VAGEN & 70.8& 63.6& 61.1& 60.0& 50.4& 61.2& 3.56&3.59& 3.67& 2.54& 2.73& 3.22& \textbf{0.64}& 0.63& \textbf{0.59}& 0.58& 0.68& 0.62\\
\midrule
\textbf{ORCA} & \textbf{73.8}& 58.4& \textbf{80.4}& \textbf{81.5}& \textbf{61.0}& \textbf{71.0}& 3.53& \textbf{3.68}& \textbf{3.93}& \textbf{3.77}& \textbf{3.67}& \textbf{3.72}& \textbf{0.64}& 0.70& 0.54& 0.63& \textbf{0.70}& \textbf{0.64}\\
\bottomrule
\end{tabular}

\vspace{0.5em}
\textbf{(b) Video Generation Quality and Human Preference}
\vspace{0.2em}

\small
\setlength{\tabcolsep}{3.5pt}
\begin{tabular}{l|ccccc|c||ccccc|c||ccccc|c}
\toprule
& \multicolumn{6}{c||}{Aesthetics $\uparrow$} & \multicolumn{6}{c||}{Subject Consistency$\uparrow$} & \multicolumn{6}{c}{BWS (\%) $\downarrow$} \\
\cmidrule{2-7} \cmidrule{8-13} \cmidrule{14-19}
Method & Kit. & Live. & Work. & Gard. & Off. & Avg & Kit. & Live. & Work. & Gard. & Off. & Avg & Kit. & Live. & Work. & Gard. & Off. & Avg \\
\midrule
Reactive & 0.61& \textbf{0.53}& 0.61& \textbf{0.62}& \textbf{0.52}&\textbf{0.59}& \textbf{0.92}& 0.93& 0.91& 0.91& \textbf{0.93}& 0.92& 0.00& -20.6& -45.0& 15.4& -40.0& -18.0\\
Open-Loop & 0.59& 0.51& 0.58& 0.60& 0.50& 0.56& 0.91& 0.91& 0.89& 0.88& 0.91& 0.90& 3.30& \textbf{8.80}& -13.3& -23.1& -13.3& -7.52\\
VAGEN & 0.60& 0.52& 0.59& 0.62& 0.50& 0.57& 0.91&0.93& 0.92& 0.91& 0.92& 0.92& -10.0& 5.90& 13.3& -23.1& -6.70& -4.12\\
\midrule
\textbf{ORCA} & \textbf{0.63}& \textbf{0.53}& \textbf{0.62}& 0.60& 0.51&0.58&\textbf{0.92}& \textbf{0.94}&\textbf{0.94}& \textbf{0.91}& 0.92& \textbf{0.93}& \textbf{6.70}& 5.90& \textbf{40.0}& \textbf{30.8}& \textbf{60.0}& \textbf{28.7}\\
\bottomrule
\end{tabular}

\vspace{0.2em}
{\footnotesize Kit.: Kitchen, Live.: Livestream, Work.: Workshop, Gard.: Garden, Off.: Office. PPS and AFS measure execution quality.}
\end{table*}

\subsection{Benchmark and Dataset}
Since L-IVA represents a novel task without direct prior work, we evaluate three representative paradigms adapted to our setting: (1) {Open-Loop Planner} plans the complete action sequence upfront given the initial scene and intention, then executes without feedback; (2) Reactive Agent~\cite{yao2023reactsynergizingreasoningacting} operates in a turn-by-turn observe-act loop without maintaining belief state and reflection; 
(3) VAGEN-style CoT~\cite{wang2025vagen} adapts world model-based 
reasoning with state estimation and transition prediction, but assumes 
deterministic environments where action outcomes are predictable.

We evaluate all methods on L-IVA across 
5 scenarios (Kitchen, Livestream, Workshop, Garden, Office) using 
multiple metrics: Task Success Rate (TSR) measuring goal completion, 
Physical Plausibility Score (PPS) and Action Fidelity Score (AFS) 
assessing execution quality, video generation quality (Aesthetics and 
Subject consistency), and Human Preference via BWS. Implementation 
details and evaluation protocols are provided in Appendix section~\ref{sec:eval_protocols}.
\subsection{Implement Details}
ORCA is training-free and leverages pre-trained models, including Gemini-2.5-Flash as the vision-language model for both System 1 and System 2, and Wanx2.2~\cite{wan2025wanopenadvancedlargescale} with distilled LoRA as the I2V generation model~\cite{lightx2v}. All baseline methods share the same VLM and generation settings to ensure fair comparison. We carefully design prompts tailored to each method's reasoning structure, with detailed prompts provided in the Appendix section~\ref{sec:prompts} .
\subsection{Comparison with Planning methods}
\subsubsection{Quantitative Comparison}
Table~\ref{tab:main_results} presents comprehensive evaluation across 5 scenarios and multiple dimensions. Results reveal important trade-offs between task completion, execution quality, and video coherence.

\noindent\textbf{Evaluation on Task Success Rate and Execution Quality}
Table~\ref{tab:main_results} presents a comprehensive evaluation across 5 scenarios. ORCA achieves the highest average Task Success Rate (71.0\%) and Physical Plausibility (3.72), validating the effectiveness of our closed-loop architecture. However, a nuanced analysis reveals an important trade-off. Open-Loop Planner remains highly competitive in scenarios with lower state dependency, such as Livestream and Kitchen, where it performs on par with or even surpasses ORCA. This phenomenon has an explanation that Open-Loop Planner schedules the complete action sequence upfront and executes all steps regardless of intermediate outcomes. In loose, low-dependency tasks, this strategy ensures the agent attempts all sub-goals within the fixed step budget, often achieving nominal completion. In contrast, ORCA’s rigorous Reflect mechanism invests computational steps in error correction; while this ensures high fidelity, it risks exhausting the step budget on retries in simpler tasks. However, the advantage of ORCA becomes decisive in complex, high-dependency environments like Garden and Workshop. In the Garden scenario, Open-Loop planner fails significantly because execution errors in early steps go undetected, rendering subsequent actions meaningless. Regarding execution quality, ORCA achieves the highest Physical Plausibility Score, significantly outperforming the Reactive Agent. The low score of the Reactive agent baseline stems from its lack of a world model; without maintaining a structured belief state, it suffers from severe object permanence issues.

\noindent\textbf{Evaluation on Video Quality and Human Preference}
Table~\ref{tab:main_results} (b) exposes a critical flaw in Open-Loop Planning, despite competitive TSR, it suffers from the lowest Subject Consistency. This validates our hypothesis that without outcome verification, visual artifacts accumulate, degrading avatar identity across long horizons. In contrast, ORCA achieves the highest Subject Consistency  by actively filtering low-quality generations in the Reflect stage. Crucially, human evaluation via Best-Worst Scaling ranks ORCA significantly higher than all baselines, with Reactive agent and Open-Loop planner receiving negative scores.
The results demonstrate that active intelligence requires more than just per-clip image quality. While Reactive agent maintains high aesthetics, its inability to execute coherent tasks leads to the lowest human preference. Open-Loop achieves task completion but sacrifices visual consistency. ORCA delivers the most robust performance, combining superior consistency with high task success, thereby validating the necessity of closed-loop world modeling.
\subsubsection{Qualitative Comparison}
Figure~\ref{fig:comapfig} compares our method with baselines on the multi-step Transfer Plant task, which requires coordinating four sequential subgoals while maintaining world state consistency.
Open-Loop Planner generates all I2V captions for the entire task sequence in one shot and directly passes them to the video generation model for execution, without any intermediate validation or reflection mechanisms. As shown in Figure~\ref{fig:comapfig} (second row), while the initial steps appear plausible, inevitable execution deviations in the generated videos (e.g., incorrect seedling removal in Step 2) go undetected and accumulate through subsequent steps. By Step 4, the generated video depicts actions on completely misaligned object.
Reactive agent enables closed-loop correction but lacks world state modeling. Row 3 shows the method repeatedly performing the same action (adding soil) without recognizing subgoal completion, resulting in physically implausible repetitive behaviors.
VAGEN combines planning with closed-loop execution but suffers from I2V model hallucinations. Without reflection mechanisms, these errors can corrupt the final state(row 4, red boxes).
ORCA successfully executes most subgoals through the OTAR cycle. The reflect phase detects errors early. This prevents all three failure modes: undetected errors, repetitive actions, and hallucination corruption. The results demonstrate that dual-system reasoning with reflection is essential for long-horizon video generation.

\subsection{Ablation Study}
\begin{table}[!ht]  
\centering
\caption{Ablation on ORCA components. Note: To ensure fair comparison within ablation variants, we re-evaluated the Workshop scene. Minor variations in absolute scores compared to Table 1 are due to the stochastic nature of human evaluation.}
\label{tab:ablation_main}
\small
\setlength{\tabcolsep}{4pt}
\begin{tabular}{lccc}
\toprule
\textbf{Variant} & \textbf{TSR↑} & \textbf{Cons.↑} & \textbf{BWS↓}\\
\midrule
ORCA (Full) & \textbf{0.77}& \textbf{0.94}& \textbf{26.7\%}\\
w/o System 1 &  0.74&  0.93&-6.72\%\\
w/o Reflect & 0.72&  0.92&  -20.0\%\\
w/o Belief State &  0.67&  0.93& 0.00\%\\
\bottomrule
\end{tabular}
\end{table}

Our ablation study systematically validates ORCA's design principles by answering three core questions established in Section~\ref{sec:intro}. 
Each question corresponds to a key challenge in realizing IWMs for stochastic generative environments.
We report metric on workshop scene.

\noindent\textbf{Is explicit world modeling necessary for active intelligence?}
Removing belief state tracking (w/o Belief State) causes the severe TSR degradation. Without maintaining $\hat{s}_t$ containing scene state, the agent cannot track completed sub-goals or reason about action dependencies, leading to repetitive or out-of-order actions.

\noindent\textbf{Is closed-loop verification necessary under generative stochasticity?}
Removing reflection (w/o Reflect) degrades Subject Consistency and BWS. Without outcome verification, incorrect generations corrupt subsequent steps. 

\noindent\textbf{Is hierarchical action specification necessary for open-domain control?}
Removing System 1's detailed grounding (w/o System 1) reduces TSR and BWS. Direct use of System 2's abstract commands yields imprecise I2V generation. This confirms that separating strategic reasoning from execution grounding is critical for reliable control.

\section{Conclusion}
This paper introduces a paradigm shift from passive animation to active intelligence in video avatars. We presented L-IVA, the first benchmark for autonomous goal completion in interactive scenarios, and ORCA, a framework grounded in Internal World Model theory. ORCA achieves robust long-horizon behavior through a closed-loop OTAR cycle for state tracking under generative uncertainty and a dual-system architecture for hierarchical planning and execution. Experimental results validate that our reflection mechanism prevents belief corruption and enables coherent multi-step task completion.

{
    \small
    \bibliographystyle{ieeenat_fullname}
    \bibliography{main}

@String(CVPR= {IEEE Conf. Comput. Vis. Pattern Recog.})

@String(CVPR  = {CVPR})

@article{xie2024dreamfactory,
  title={Dreamfactory: Pioneering multi-scene long video generation with a multi-agent framework},
  author={Xie, Zhifei and Tang, Daniel and Tan, Dingwei and Klein, Jacques and Bissyand, Tegawend F and Ezzini, Saad},
  journal={arXiv preprint arXiv:2408.11788},
  year={2024}
}

@article{hu2024storyagent,
  title={Storyagent: Customized storytelling video generation via multi-agent collaboration},
  author={Hu, Panwen and Jiang, Jin and Chen, Jianqi and Han, Mingfei and Liao, Shengcai and Chang, Xiaojun and Liang, Xiaodan},
  journal={arXiv preprint arXiv:2411.04925},
  year={2024}
}

@article{yuan2024mora,
  title={Mora: Enabling generalist video generation via a multi-agent framework},
  author={Yuan, Zhengqing and Liu, Yixin and Cao, Yihan and Sun, Weixiang and Jia, Haolong and Chen, Ruoxi and Li, Zhaoxu and Lin, Bin and Yuan, Li and He, Lifang and others},
  journal={arXiv preprint arXiv:2403.13248},
  year={2024}
}

@article{zheng2024videogen,
  title={Videogen-of-thought: A collaborative framework for multi-shot video generation},
  author={Zheng, Mingzhe and Xu, Yongqi and Huang, Haojian and Ma, Xuran and Liu, Yexin and Shu, Wenjie and Pang, Yatian and Tang, Feilong and Chen, Qifeng and Yang, Harry and others},
  journal={arXiv e-prints},
  pages={arXiv--2412},
  year={2024}
}

@article{long2025vista,
  title={VISTA: A Test-Time Self-Improving Video Generation Agent},
  author={Long, Do Xuan and Wan, Xingchen and Nakhost, Hootan and Lee, Chen-Yu and Pfister, Tomas and Ar{\i}k, Sercan {\"O}},
  journal={arXiv preprint arXiv:2510.15831},
  year={2025}
}

@article{huang2024genmac,
  title={Genmac: compositional text-to-video generation with multi-agent collaboration},
  author={Huang, Kaiyi and Huang, Yukun and Ning, Xuefei and Lin, Zinan and Wang, Yu and Liu, Xihui},
  journal={arXiv preprint arXiv:2412.04440},
  year={2024}
}

@article{ma2025sayanything,
  title={Sayanything: Audio-driven lip synchronization with conditional video diffusion},
  author={Ma, Junxian and Wang, Shiwen and Yang, Jian and Hu, Junyi and Liang, Jian and Lin, Guosheng and Li, Kai and Meng, Yu and others},
  journal={arXiv preprint arXiv:2502.11515},
  year={2025}
}

@inproceedings{prajwal2020lip,
  title={A lip sync expert is all you need for speech to lip generation in the wild},
  author={Prajwal, KR and Mukhopadhyay, Rudrabha and Namboodiri, Vinay P and Jawahar, CV},
  booktitle={Proceedings of the 28th ACM international conference on multimedia},
  pages={484--492},
  year={2020}
}

@inproceedings{lin2025omnihuman,
  title={Omnihuman-1: Rethinking the scaling-up of one-stage conditioned human animation models},
  author={Lin, Gaojie and Jiang, Jianwen and Yang, Jiaqi and Zheng, Zerong and Liang, Chao and Zhang, Yuan and Liu, Jingtuo},
  booktitle={Proceedings of the IEEE/CVF International Conference on Computer Vision},
  pages={13847--13858},
  year={2025}
}

@article{jiang2025omnihuman,
  title={Omnihuman-1.5: Instilling an active mind in avatars via cognitive simulation},
  author={Jiang, Jianwen and Zeng, Weihong and Zheng, Zerong and Yang, Jiaqi and Liang, Chao and Liao, Wang and Liang, Han and Zhang, Yuan and Gao, Mingyuan},
  journal={arXiv preprint arXiv:2508.19209},
  year={2025}
}

@article{he2024id,
  title={Id-animator: Zero-shot identity-preserving human video generation},
  author={He, Xuanhua and Liu, Quande and Qian, Shengju and Wang, Xin and Hu, Tao and Cao, Ke and Yan, Keyu and Zhang, Jie},
  journal={arXiv preprint arXiv:2404.15275},
  year={2024}
}

@inproceedings{wang2025dynamicface,
  title={Dynamicface: High-quality and consistent face swapping for image and video using composable 3d facial priors},
  author={Wang, Runqi and Chen, Yang and Xu, Sijie and He, Tianyao and Zhu, Wei and Song, Dejia and Chen, Nemo and Tang, Xu and Hu, Yao},
  booktitle={Proceedings of the IEEE/CVF International Conference on Computer Vision},
  pages={13438--13447},
  year={2025}
}

@article{zhong2025concat,
  title={Concat-ID: Towards Universal Identity-Preserving Video Synthesis},
  author={Zhong, Yong and Yang, Zhuoyi and Teng, Jiayan and Gu, Xiaotao and Li, Chongxuan},
  journal={arXiv preprint arXiv:2503.14151},
  year={2025}
}

@article{huang2024consistentid,
  title={Consistentid: Portrait generation with multimodal fine-grained identity preserving},
  author={Huang, Jiehui and Dong, Xiao and Song, Wenhui and Chong, Zheng and Tang, Zhenchao and Zhou, Jun and Cheng, Yuhao and Chen, Long and Li, Hanhui and Yan, Yiqiang and others},
  journal={arXiv preprint arXiv:2404.16771},
  year={2024}
}

@article{wang2025interacthuman,
  title={InterActHuman: Multi-Concept Human Animation with Layout-Aligned Audio Conditions},
  author={Wang, Zhenzhi and Yang, Jiaqi and Jiang, Jianwen and Liang, Chao and Lin, Gaojie and Zheng, Zerong and Yang, Ceyuan and Lin, Dahua},
  journal={arXiv preprint arXiv:2506.09984},
  year={2025}
}

@article{Yang2025InfiniteTalkAV,
  title={InfiniteTalk: Audio-driven Video Generation for Sparse-Frame Video Dubbing},
  author={Shaoshu Yang and Zhe Kong and Feng Gao and Meng Cheng and Xiangyu Liu and Yong Zhang and Zhuoliang Kang and Wenhan Luo and Xunliang Cai and Ran He and Xiaoming Wei},
  journal={ArXiv},
  year={2025},
  volume={abs/2508.14033},
  url={https://api.semanticscholar.org/CorpusID:280686279}
}

@inproceedings{Tian2024EMOEP,
  title={EMO: Emote Portrait Alive - Generating Expressive Portrait Videos with Audio2Video Diffusion Model under Weak Conditions},
  author={Linrui Tian and Qi Wang and Bang Zhang and Liefeng Bo},
  booktitle={European Conference on Computer Vision},
  year={2024},
  url={https://api.semanticscholar.org/CorpusID:268032834}
}

@article{Tu2024StableAnimatorHI,
  title={StableAnimator: High-Quality Identity-Preserving Human Image Animation},
  author={Shuyuan Tu and Zhen Xing and Xintong Han and Zhi-Qi Cheng and Qi Dai and Chong Luo and Zuxuan Wu},
  journal={2025 IEEE/CVF Conference on Computer Vision and Pattern Recognition (CVPR)},
  year={2024},
  pages={21096-21106},
  url={https://api.semanticscholar.org/CorpusID:274281130}
}

@article{Gan2025OmniAvatarEA,
  title={OmniAvatar: Efficient Audio-Driven Avatar Video Generation with Adaptive Body Animation},
  author={Qijun Gan and Ruizi Yang and Jianke Zhu and Shaofei Xue and Steven Hoi},
  journal={ArXiv},
  year={2025},
  volume={abs/2506.18866},
  url={https://api.semanticscholar.org/CorpusID:280000803}
}

@article{Hu2025AnimateA2,
  title={Animate Anyone 2: High-Fidelity Character Image Animation with Environment Affordance},
  author={Liucheng Hu and Guangyuan Wang and Zhen Shen and Xin Gao and Dechao Meng and Lian Zhuo and Peng Zhang and Bang Zhang and Liefeng Bo},
  journal={ArXiv},
  year={2025},
  volume={abs/2502.06145},
  url={https://api.semanticscholar.org/CorpusID:276249182}
}

@article{Ma2023FollowYP,
  title={Follow Your Pose: Pose-Guided Text-to-Video Generation using Pose-Free Videos},
  author={Yue Ma and Yin-Yin He and Xiaodong Cun and Xintao Wang and Ying Shan and Xiu Li and Qifeng Chen},
  journal={ArXiv},
  year={2023},
  volume={abs/2304.01186},
  url={https://api.semanticscholar.org/CorpusID:257912672}
}

@incollection{spaan2012partially,
  title={Partially observable Markov decision processes},
  author={Spaan, Matthijs TJ},
  booktitle={Reinforcement learning: State-of-the-art},
  pages={387--414},
  year={2012},
  publisher={Springer}
}

@article{wang2025vagen,
  title={VAGEN: Reinforcing world model reasoning for multi-turn vlm agents},
  author={Wang, Kangrui and Zhang, Pingyue and Wang, Zihan and Gao, Yaning and Li, Linjie and Wang, Qineng and Chen, Hanyang and Wan, Chi and Lu, Yiping and Yang, Zhengyuan and others},
  journal={arXiv preprint arXiv:2510.16907},
  year={2025}
}

@article{diester2024internal,
  title={Internal world models in humans, animals, and AI},
  author={Diester, Ilka and Bartos, Marlene and B{\"o}decker, Joschka and Kortylewski, Adam and Leibold, Christian and Letzkus, Johannes and Nour, Matthew M and Sch{\"o}nauer, Monika and Straw, Andrew and Valada, Abhinav and others},
  journal={Neuron},
  volume={112},
  number={14},
  pages={2265--2268},
  year={2024},
  publisher={Elsevier}
}

@article{chen2025internalizing,
  title={Internalizing World Models via Self-Play Finetuning for Agentic RL},
  author={Chen, Shiqi and Zhu, Tongyao and Wang, Zian and Zhang, Jinghan and Wang, Kangrui and Gao, Siyang and Xiao, Teng and Teh, Yee Whye and He, Junxian and Li, Manling},
  journal={arXiv preprint arXiv:2510.15047},
  year={2025}
}

@article{evans2013dual,
  title={Dual-process theories of higher cognition: Advancing the debate},
  author={Evans, Jonathan St BT and Stanovich, Keith E},
  journal={Perspectives on psychological science},
  volume={8},
  number={3},
  pages={223--241},
  year={2013},
  publisher={Sage Publications Sage CA: Los Angeles, CA}
}

@article{gao2025vision,
  title={Do vision-language models have internal world models? towards an atomic evaluation},
  author={Gao, Qiyue and Pi, Xinyu and Liu, Kevin and Chen, Junrong and Yang, Ruolan and Huang, Xinqi and Fang, Xinyu and Sun, Lu and Kishore, Gautham and Ai, Bo and others},
  journal={arXiv preprint arXiv:2506.21876},
  year={2025}
}

@article{cen2025worldvla,
  title={WorldVLA: Towards Autoregressive Action World Model},
  author={Cen, Jun and Yu, Chaohui and Yuan, Hangjie and Jiang, Yuming and Huang, Siteng and Guo, Jiayan and Li, Xin and Song, Yibing and Luo, Hao and Wang, Fan and others},
  journal={arXiv preprint arXiv:2506.21539},
  year={2025}
}

@misc{yao2023reactsynergizingreasoningacting,
      title={ReAct: Synergizing Reasoning and Acting in Language Models}, 
      author={Shunyu Yao and Jeffrey Zhao and Dian Yu and Nan Du and Izhak Shafran and Karthik Narasimhan and Yuan Cao},
      year={2023},
      eprint={2210.03629},
      archivePrefix={arXiv},
      primaryClass={cs.CL},
      url={https://arxiv.org/abs/2210.03629}, 
}

@book{louviere2015best,
  title={Best-worst scaling: Theory, methods and applications},
  author={Louviere, Jordan J and Flynn, Terry N and Marley, Anthony Alfred John},
  year={2015},
  publisher={Cambridge University Press}
}

@misc{lightx2v,
 author = {LightX2V Contributors},
 title = {LightX2V: Light Video Generation Inference Framework},
 year = {2025},
 publisher = {GitHub},
 journal = {GitHub repository},
 howpublished = {\url{https://github.com/ModelTC/lightx2v}},
}

@misc{wan2025wanopenadvancedlargescale,
      title={Wan: Open and Advanced Large-Scale Video Generative Models}, 
      author={Team Wan and Ang Wang and Baole Ai and Bin Wen and Chaojie Mao and Chen-Wei Xie and Di Chen and Feiwu Yu and Haiming Zhao and Jianxiao Yang and Jianyuan Zeng and Jiayu Wang and Jingfeng Zhang and Jingren Zhou and Jinkai Wang and Jixuan Chen and Kai Zhu and Kang Zhao and Keyu Yan and Lianghua Huang and Mengyang Feng and Ningyi Zhang and Pandeng Li and Pingyu Wu and Ruihang Chu and Ruili Feng and Shiwei Zhang and Siyang Sun and Tao Fang and Tianxing Wang and Tianyi Gui and Tingyu Weng and Tong Shen and Wei Lin and Wei Wang and Wei Wang and Wenmeng Zhou and Wente Wang and Wenting Shen and Wenyuan Yu and Xianzhong Shi and Xiaoming Huang and Xin Xu and Yan Kou and Yangyu Lv and Yifei Li and Yijing Liu and Yiming Wang and Yingya Zhang and Yitong Huang and Yong Li and You Wu and Yu Liu and Yulin Pan and Yun Zheng and Yuntao Hong and Yupeng Shi and Yutong Feng and Zeyinzi Jiang and Zhen Han and Zhi-Fan Wu and Ziyu Liu},
      year={2025},
      eprint={2503.20314},
      archivePrefix={arXiv},
      primaryClass={cs.CV},
      url={https://arxiv.org/abs/2503.20314}, 
}
}

\clearpage
\setcounter{page}{1}
\maketitlesupplementary

\section{Supplementary Material}
This supplementary material provides comprehensive details to support the main paper. It is organized as follows: Section A details the construction of the L-IVA benchmark; Section B elaborates on the evaluation protocols and metrics; Section C presents additional qualitative visualizations comparing different methods; Section D lists the specific prompts utilized in the ORCA framework; Section E discuss the failure case and limitations.
\section{Benchmark Construction Details}
\label{sec:benchmark_details}
To ensure diversity in visual complexity and environmental dynamics, we construct the L-IVA benchmark through a hybrid pipeline combining real-world photography and AI-synthesized imagery.

\begin{figure*}
    \centering
    \includegraphics[width=\linewidth]{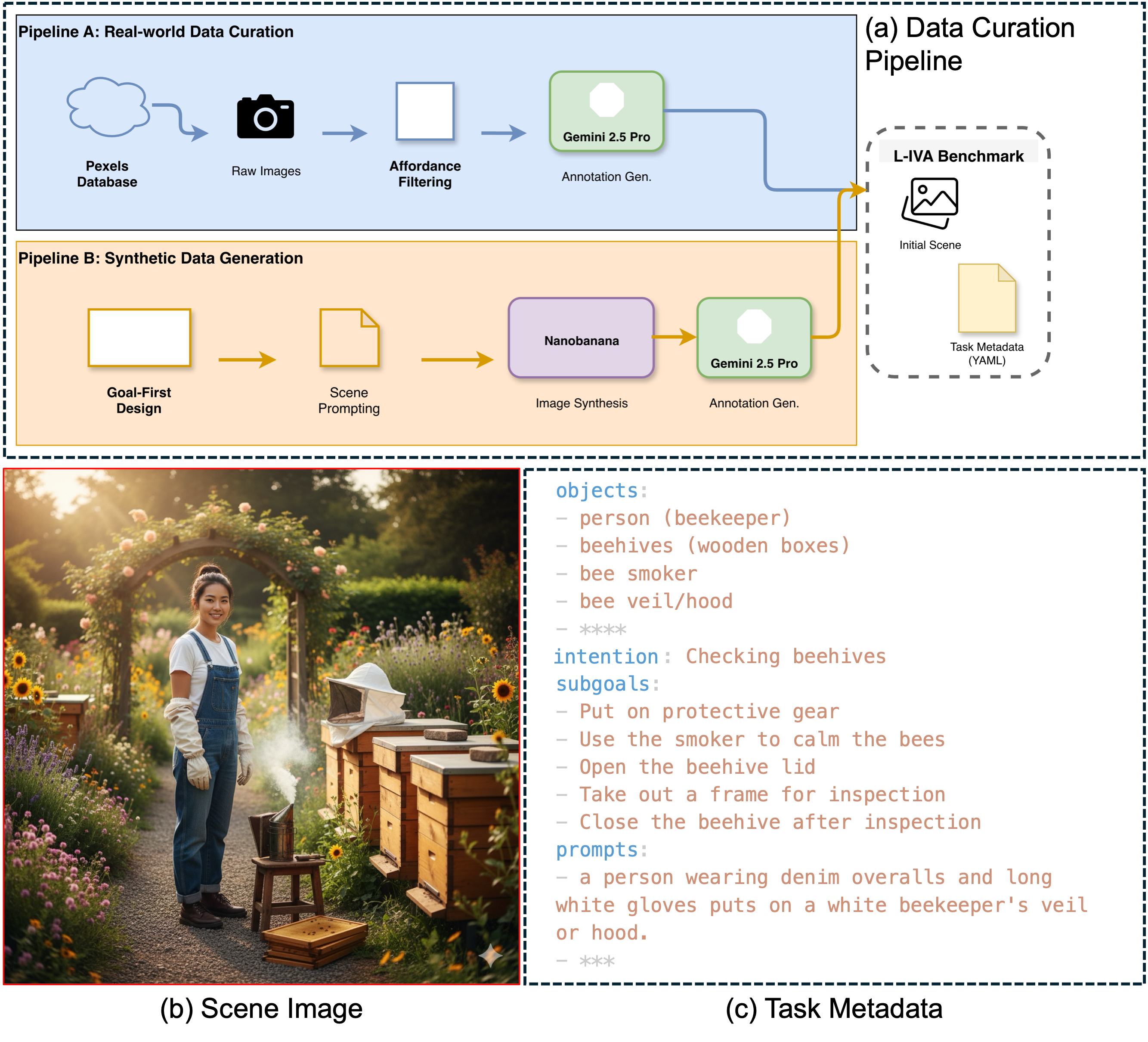}
  \caption{\textbf{Overview of the L-IVA Benchmark Construction Pipeline.} (a) Our data curation process employs a hybrid strategy: \textbf{Pipeline A} sources real-world images from Pexels, filtered by scene affordance and annotated via Gemini-2.5-Pro. \textbf{Pipeline B} utilizes a goal-first design for synthetic data, where scenes are generated by Nanobanana to strictly align with intended interactions. (b) A representative scene image (e.g., "Checking beehives") from the benchmark. (c) The corresponding structured metadata (YAML), including object inventory, high-level intention, subgoals, and reference prompts.}
  \label{fig:benchmark_pipeline}
\end{figure*}

\paragraph{Real-world Data Curation.}
We source high-quality real-world images from Pexels to serve as initial observations. The selection process is strictly guided by \textit{scene affordance}: we filter for images containing distinct, interactive objects capable of supporting multi-step physical manipulations necessary to achieve high-level goals. For each selected scene, we manually define a high-level intention ($I$). To generate ground-truth annotations efficiently, we leverage Gemini-2.5-Pro. Given the image and the intention, the model generates a structured set of metadata, including sequential subgoals, detailed object descriptions, and reference action prompts. Each data sample is stored as a pair consisting of the initial image and a corresponding YAML file containing these hierarchical annotations (see Figure~\ref{fig:benchmark_pipeline}(c)).

\paragraph{Synthetic Data Generation.}
For synthetic scenarios, we utilize Nanobanana to create controlled environments with specific object configurations. Unlike the real-world pipeline, we adopt a \textit{goal-first approach}: we first design a high-level intention and ensure that all requisite object interactions are logically solvable within a single scene. Based on these requirements, we craft detailed text prompts to generate the initial scene image. This ``design-then-generate'' strategy ensures precise alignment between the visual assets (objects in the scene) and the task requirements, guaranteeing that the generated environments inherently support the intended interaction sequence.
\section{Evaluation Protocols}
\label{sec:eval_protocols}
\begin{figure*}
        \centering
    \includegraphics[width=\linewidth]{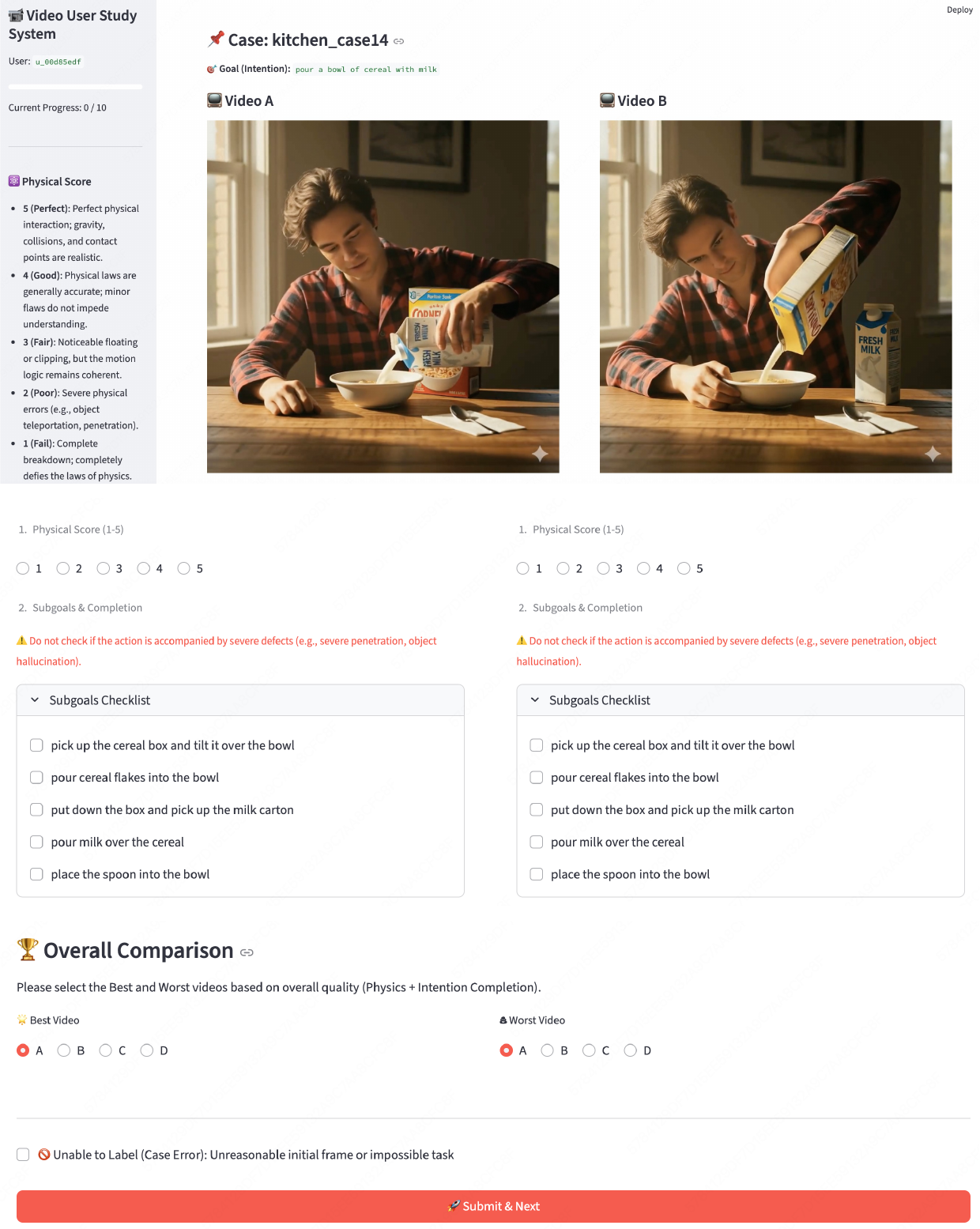}
    \caption{The user interface for human evaluation. For each test case, annotators are presented with a high-level goal (Intention) and generated videos from four anonymized methods (Video A and Video B are shown here; C and D are omitted for brevity). Evaluators are asked to assess each video based on two criteria: (1) a Physical Score (1-5 Likert scale) regarding simulation stability, and (2) a Subgoals Checklist to verify task completion. Finally, they select the best and worst videos based on overall quality.
}
    \label{fig:user_study}

\end{figure*}

To systematically assess the capabilities of active video avatars, we employ a hybrid evaluation strategy combining VLM-based automated metrics and human judgment. All automated evaluations utilize \texttt{Gemini-2.5-Flash} due to its strong multimodal reasoning and long-context video understanding capabilities. We use user study to measure the physical score and TSR, as current VLMs struggle with subtle physical inconsistencies and long-horizon causal reasoning. The webpage of the user study is shown in Figure~\ref{fig:user_study}. Our user study involves 8 human evaluators. We constructed 130 comparative evaluation sets sampled across different scenarios. Since each set includes videos from 4 methods (ORCA + 3 baselines), this results in a total of 520 evaluated videos, ensuring statistical reliability. The evaluation focuses on three key dimensions:
\paragraph{Task Success Rate (TSR).}
TSR measures the agent's ability to complete the high-level intention through multi-step subgoals. Unlike single-clip generation, our task requires causal completion of a sequence of actions. Human evaluators are presented with the high-level intention $I$, the ground-truth subgoal list $\{g_1, ..., g_N\}$, and the generated video sequence $V$. For each sample, evaluators count the number of \textit{successfully completed subgoals} based on visual evidence.
The final TSR score is calculated as the ratio of completed subgoals to the total number of subgoals:
\begin{equation}
    \text{TSR} = \frac{1}{K} \sum_{k=1}^{K} \frac{N_{\text{completed}}^{(k)}}{N_{\text{total}}^{(k)}}
\end{equation}
where $N_{\text{completed}}^{(k)}$ is the number of subgoals successfully executed in the $k$-th test case.

\paragraph{Action Fidelity Score (AFS).\label{afs}} AFS measures the semantic alignment between the planned action command $a_t$ and the executed video clip $v_t$. As detailed in \textbf{Table~\ref{tab:prompt_afs}}, this is a binary classification task ($0/1$). The VLM verifies if the \textit{Core Action}, \textit{Key Objects}, and \textit{Directional Consistency} in the video match the text caption. It is designed to be tolerant of minor visual artifacts but strict on semantic errors (e.g., performing "cut" instead of "peel").
\paragraph{Physical Plausibility Score (PPS)\label{pps}.}
This metric evaluates the physical consistency of the generated world, specifically targeting I2V-specific hallucinations. Human evaluators score each video sequence on a \textbf{1-5 Likert scale}:
\begin{itemize}
    \item \textbf{5 (Perfect):} Perfect physical interaction with realistic gravity, collision, and contact points.
    \item \textbf{4 (Good):} Physics generally correct; minor artifacts do not impede comprehension.
    \item \textbf{3 (Fair):} Noticeable floating or clipping, but the action logic remains coherent.
    \item \textbf{2 (Poor):} Severe physical violations (e.g., object teleportation, interpenetration).
    \item \textbf{1 (Fail):} Complete breakdown; failure to adhere to basic physical laws.
\end{itemize}
\paragraph{Human Preference Ranking Protocol.}
To evaluate holistic performance, we employ Best-Worst Scaling (BWS), which is shown to be more statistically robust than simple ranking. For each query, annotators are presented with the initial scene, the intention, and anonymized videos generated by the comparing methods side-by-side.
Annotators are instructed to select the \textbf{Best} and \textbf{Worst} models based on a hierarchical criterion: first prioritizing \textit{Task Completion} (whether the intention was fulfilled), followed by \textit{Logical Coherence} (smoothness of the action plan), and finally \textit{Visual Aesthetics}.
The BWS score for each method is computed as:
\begin{equation}
    \text{BWS Score} = \frac{\#\text{Best} - \#\text{Worst}}{\#\text{Total Cases}} \times 100\%
\end{equation}
This score ranges from -100\% to +100\%, where a positive score indicates that the method was chosen as "Best" more often than "Worst".
\section{Prompt Details for ORCA}
\label{sec:prompts}

\begin{figure*}
    \centering
    \includegraphics[width=\linewidth]{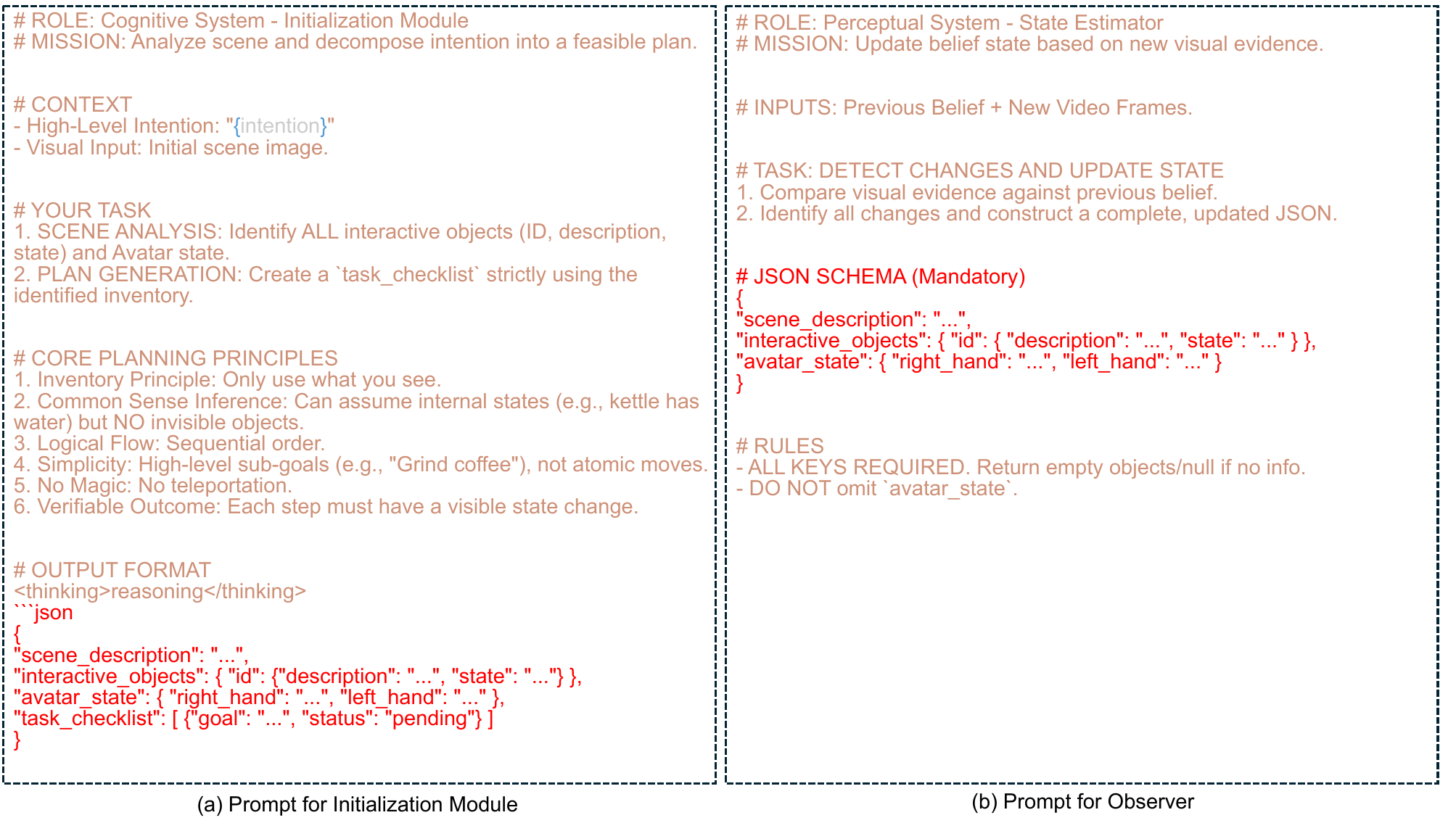}
    \caption{Prompt for Initialization module and observer}
    \label{fig:prompt1}
\end{figure*}

\begin{figure*}
    \centering
    \includegraphics[width=\linewidth]{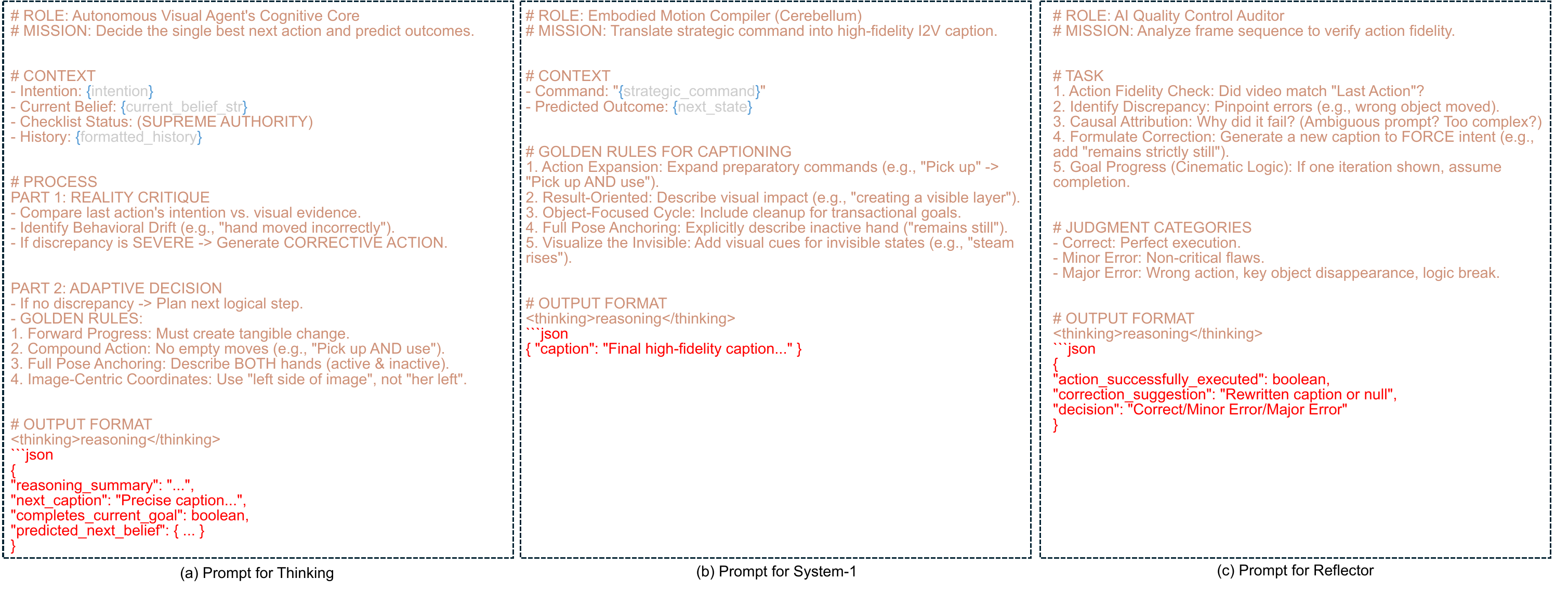}
    \caption{Prompt for Thinking, System-1 and Reflector}
    \label{fig:prompt2}
\end{figure*}

In this section, we provide the full system prompts used in the ORCA framework. These prompts are designed to be model-agnostic but are optimized for Gemini-2.5-Flash. 
\begin{itemize}
    \item \textbf{Figure~\ref{fig:prompt1}}(a): The \textit{Initialization Prompt} used by System 2 to parse the initial scene and decompose the high-level intention into a structured plan.
    \item \textbf{Figure~\ref{fig:prompt1}}(b): The \textit{Observation Prompt} for updating the belief state based on generated video clips.
    \item \textbf{Figure~\ref{fig:prompt2}}(a): The \textit{Thinking Prompt} (System 2) for strategic reasoning, discrepancy detection, and next-step planning.
    \item \textbf{Figure~\ref{fig:prompt2}}(b): The \textit{Action Grounding Prompt} (System 1) for translating abstract plans into high-fidelity, I2V-compatible captions.
    \item \textbf{Figure~\ref{fig:prompt2}}(c): The \textit{Reflection Prompt} for verifying action execution and triggering error correction.
\end{itemize}

\section{Discussion on Failure Cases and Limitations}
\label{sec:failure_analysis}

\begin{figure*}
    \centering
    \includegraphics[width=\linewidth]{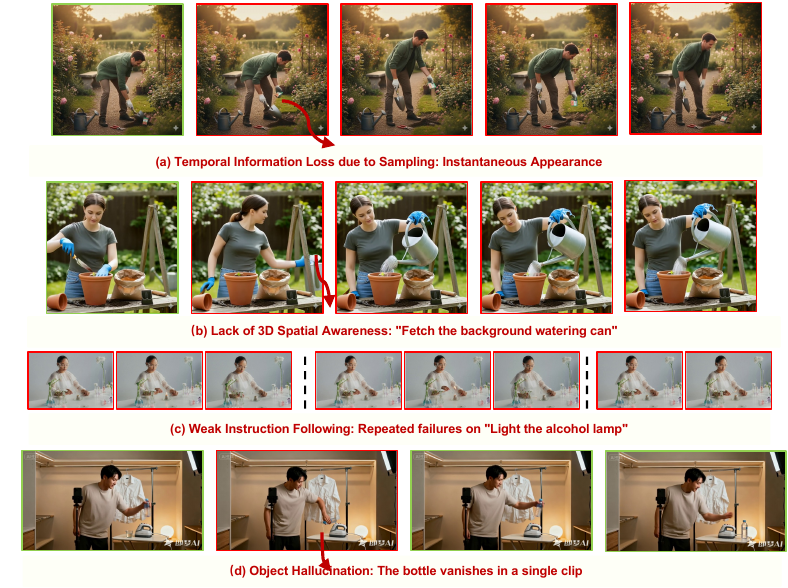}
  \caption{\textbf{Qualitative Analysis of Failure Cases attributed to Foundation Model Limitations.} 
  (a) \textbf{Temporal Information Loss:} Due to discrete frame sampling, the VLM misses the "teleportation" artifact where the fertilizer bag instantaneously appears in the hand (red arrow), falsely accepting it as a valid pickup.
  (b) \textbf{Lack of 3D Spatial Awareness:} The VLM misinterprets the depth of the scene, instructing the avatar to fetch a watering can that is actually in the distant background, resulting in an unnatural reaching motion.
  (c) \textbf{Weak Instruction Following:} For fine-grained tasks like "light the alcohol lamp," the I2V model consistently fails to execute the interaction despite ORCA triggering multiple retries (separated by dashed lines).
  (d) \textbf{Object Disappearance:} A clear example of generative instability where a key object (the water bottle) vanishes mid-clip (red box) despite no interaction occurring.}
  \label{fig:failure_cases}
\end{figure*}

While ORCA demonstrates superior performance in long-horizon task execution compared to open-loop baselines, its capabilities are inevitably bounded by the underlying foundation models (VLM and I2V). We categorize the observed failure cases into two dimensions: \textit{Perceptual Bottlenecks} (VLM-side) and \textit{Generative Constraints} (I2V-side). It is important to note that these failures stem primarily from the intrinsic limitations of current pre-trained models rather than the algorithmic design of the ORCA framework.
\paragraph{VLM-Centric Failures: Perception and Depth Ambiguity.}
The reliability of ORCA's \textit{Reflect} and \textit{Observe} modules depends on the VLM's visual grounding ability. We observe two specific issues:
\begin{itemize}
    \item \textbf{Temporal Information Loss due to Sampling:} To manage context length, ORCA feeds sampled frames (e.g., 5 frames) to the VLM. This discrete sampling can cause \textit{Temporal Aliasing}, where critical but fleeting glitches (e.g., an object flickering out of existence for just 2 frames) fall between sampled frames. Consequently, the VLM may generate a "False Positive" judgment, accepting a flawed video.
    \item \textbf{Lack of 3D Spatial Awareness:} Current VLMs operate in 2D pixel space and often struggle with depth perception. As illustrated in Figure~\ref{fig:failure_cases}(b), the VLM may misinterpret a background object as being within the avatar's reach, leading to geometrically impossible instructions (e.g., "pick up the distant cup") that the I2V model cannot execute realistically.
\end{itemize}
\paragraph{I2V-Centric Failures: Control and Consistency.}
Even when ORCA generates perfect instructions, the generative backbone (I2V model) acts as a physical execution bottleneck:
\begin{itemize}
    \item \textbf{Instruction Following:} In handling complex, fine-grained manipulations, the I2V model often exhibits strong prior biases and fails to adhere to the prompt. Although ORCA's \textit{Reflect} module correctly rejects these failures and triggers retries (up to $N_{retry}$ times), the I2V model may persistently fail to generate the correct physics, leading to an eventual task termination.
    \item \textbf{Object Permanence and Hallucination:} Generative models inherently struggle with long-term object permanence. As shown in Figure~\ref{fig:failure_cases}(d), sudden object disappearance or the spontaneous hallucination of new objects can occur. While ORCA's state tracking attempts to catch these errors, severe hallucinations can sometimes corrupt the agent's belief state if they are too subtle for the VLM to detect immediately.
\end{itemize}
\paragraph{Conclusion.}
These limitations highlight that ORCA is a \textit{framework for active intelligence}, currently operating on imperfect substrates. We posit that as VLM spatial reasoning and I2V controllability improve, ORCA's performance will scale accordingly without architectural changes.

\clearpage 

\begin{table*}[t]
    \centering
    \scriptsize
    \renewcommand{\arraystretch}{1.1}
    \caption{Full Prompt for Action Fidelity Score (AFS) Evaluation}
    \label{tab:prompt_afs}
    \begin{tabularx}{\linewidth}{|X|}
        \hline
        \textbf{Prompt Content} \\ 
        \hline
        \ttfamily
        \# ROLE: Action Fidelity Evaluator \par
        You are an expert evaluator assessing alignment between a video clip and its action caption. \par
        \medskip
        \#\# EVALUATION CRITERIA \par
        ALIGNED (af = 1) if: \par
        1. Core Action Match: Primary verb (e.g., "pick up") is clearly visible. \par
        2. Object Correctness: Key objects match the caption. \par
        3. Spatial/Directional Consistency: Motion direction and location match roughly. \par
        \medskip
        NOT ALIGNED (af = 0) if: \par
        1. Wrong Action: Action shown is fundamentally different (e.g., "cut" vs "stir"). \par
        2. Wrong Object: Object manipulated is incorrect. \par
        3. Missing Action: Described action is absent. \par
        4. Impossible Outcome: Visual contradicts caption's physics. \par
        \medskip
        \#\# TOLERANCE GUIDELINES \par
        - ACCEPT: Visual quality issues (blur), Incomplete visibility (partially off-screen), Semantic equivalence ("bowl" vs "container"), Timing flexibility. \par
        - REJECT: Contradictory actions, Wrong action type, Missing critical steps. \par
        \medskip
        \#\# OUTPUT FORMAT \par
        Provide reasoning in <thinking>, then output JSON: \par
        \{ \par
        \quad "af": 0 or 1, \par
        \quad "reason": "Brief explanation..." \par
        \} \\
        \hline
    \end{tabularx}
\end{table*}


\end{document}